\documentclass{article}

\usepackage{microtype}
\usepackage{graphicx}
\usepackage{subfigure}
\usepackage{booktabs} 

\usepackage{hyperref}

\usepackage[accepted]{sysml2019}

\sysmltitlerunning{Energy-aware DNN Graph Optimization}

\begin{document}

\twocolumn[
\sysmltitle{Energy-aware DNN Graph Optimization}



\sysmlsetsymbol{equal}{*}

\begin{sysmlauthorlist}
\sysmlauthor{Yu Wang}{ClemsonSOC,USTCSSE}
\sysmlauthor{Rong Ge}{ClemsonSOC}
\sysmlauthor{Shuang Qiu}{ClemsonSOC,USTCSSE}
\end{sysmlauthorlist}

\sysmlaffiliation{ClemsonSOC}{School of Computing, Clemson University, Clemson, South Carolina, USA}
\sysmlaffiliation{USTCSSE}{School of Software Engineering, University of Science and Technology of China, Hefei, Anhui, China}

\sysmlcorrespondingauthor{Rong Ge}{rge@clemson.edu}

\sysmlkeywords{Machine Learning, ReCoML}

\vskip 0.3in

\begin{abstract}
Unlike existing work in deep neural network (DNN) graphs optimization for inference performance, we explore  DNN graph optimization for energy awareness and savings for power- and resource-constrained machine learning devices. We present a method that allows users to optimize energy consumption or balance between energy and inference performance for  DNN graphs.
This method efficiently searches through the space of equivalent graphs, and  identifies a graph and the corresponding algorithms  that incur the least cost in execution. 
We implement the method and evaluate it  with multiple DNN models on a GPU-based machine. Results show that our method achieves significant  energy savings, i.e., 24\% with negligible performance impact.  
\end{abstract}
]



\printAffiliationsAndNotice{}  

\section{Introduction}
\label{introduction}

Machine Learning (ML), especially deep neural network (DNN) technologies are changing nearly every area of our lives from transportation and health care to science and security. As DNN models become deeper and more complex, their training and inference involve intensive computation  and consume more power and energy. Nevertheless, power is a key constraint on many computing platforms, especially for battery-power devices. Today smartphones and wearable devices run a large number of  ML applications, and power savings for these applications can improve the performance of other designed functionalities and prolong the device functioning time. Meanwhile, power and energy savings on datacenters can  reduce the energy bills and green gas emission. 

In this work,  we explore energy-aware graph optimization through  graph substitution. By substituting one or more nodes of a graph while maintaining the same functionalities, graph substitution yields a space of equivalent graphs. These graphs, however, incur different costs for performing the functionalities, and the ones with the least cost are optimal. DNN models can be represented as graphs, where the nodes are operators and edges are the tensors. Graph substitution has been used in DNN frameworks (\citet{abadi2016tensorflow}; \citet{paszke2017tensors}; \citet{chen2018tvm}; \citet{jia2019opt_dnn})  to optimize the performance (or delay) of training and inference. It has the advantage over model pruning in that it maintains accuracy. Nevertheless, as we demonstrate in this paper, previous work that uses substitution to optimize delay  generally doesn't lead to  optimal energy,  and sometimes increases energy for no or little delay improvement. To support DL systems and scenarios severely constrained by power and energy, we investigate graph substitution for optimal energy and energy-delay tradeoffs.  

Identifying the optimal graphs requires cost quantification for the graphs. One  quantification method is direct measurement that runs every graph of interest and records the cost. This is time-consuming and impractical for non-dedicated systems and large search spaces.
To alleviate this issue, we build analytical models that predict from graph architecture and node/edge properties the cost, i.e., inference time, energy, or power. This analytical method quickly estimates costs   and is applicable for  online employment on various systems and platforms.

The cost of a DNN graph  depends on not only its architecture but also the algorithms and implementations performing the operations on the nodes. For a given DNN graph, its operators can be performed with multiple algorithms that incur different costs. For example, existing deep learning frameworks (e.g. Tensorflow\cite{abadi2016tensorflow}; Pytorch\cite{paszke2017tensors}; Metaflow\cite{jia2019opt_dnn}) rely on the underlying cuDNN\cite{chetlur2014cudnn} library, which comprises multiple algorithms for convolution operations. 
These algorithms  consume very different amounts of energy, depending on the convolution scheme, data volume and hardware architecture. 

Table \ref{table1} presents the inference time, power and energy consumption of three different\footnote{They are different because they use different parameters including input dimension, number of output layer, kernel width \& height, stride width \& height, padding width \& height,  use activation or not.} convolution operations under different algorithms on a NVIDIA Tesla V100 GPU.
\begin{table*}[t]
	\caption{Costs of different DNN graph nodes under different algorithms. (Time stands for inference time in ms; Power stands for average power in  Watt; Energy stands for energy consumption per 1000 inference in Joule.)
}
	\label{table1}
	\begin{center}
		\begin{small}
			\begin{sc}
				\setlength{\tabcolsep}{1.5mm}{}
				\begin{tabular}{l|ccc|ccc|ccc}
					\hline
					&\multicolumn{3}{|c|}{algorithm a} &\multicolumn{3}{|c|}{algorithm b}&\multicolumn{3}{c}{algorithm c} \\
					node&time&power&energy&time&power&energy&time&power&energy \\
					\hline
					
					conv1 &0.0195&144.5&2.81&0.0209(1.07$\times$)&84&1.75(0.62$\times$)&-&-&- \\
					conv2 &0.00941&58&0.545&0.0175(1.85$\times$)&47&0.822(1.50$\times$)&-&-&-\\
					conv3 &0.165&190.8&31.4&0.146(0.88$\times$)&116&16.9(0.53$\times$)&0.083(0.50$\times$)&144&11.9(0.37$\times$) \\
					\bottomrule
				\end{tabular}
			\end{sc}
		\end{small}
	\end{center}
	\vskip -0.1in
\end{table*}
For Conv1, algorithm B is 7\% slower but  consumes 38\% less energy than algorithm A.  Algorithm C is not applicable to this specific convolution operation \footnote{Some cuDNN algorithms are not applicable to all convolution operators.}. 
For Conv2, algorithm B is slower and consumes more energy than algorithm A. 
For Conv3, algorithm C, which is applicable in this case, is the fastest and consumes the least energy. 
We should pick algorithms B, A, and C for these three convolution operations respectively to optimize energy, but A, A, and C to optimize time. 
This example demonstrates that given an optimization objective the best algorithms for different nodes can be different, and for a given node the best algorithms for different optimization objectives can also be different.

Unlike previous work considering only equivalent graphs, ours considers an expanded  search space with both equivalent graphs and  algorithm assignments for the nodes. This extension exploits the energy impacts of algorithms and achieves greater gains.  Nevertheless, it  increases the computation requirement exponentially for cost quantification. To alleviate this challenge, we propose a two-level search algorithm, which identifies the  graph with an associated algorithm assignments that best matches the cost functions in a reasonable time. 



Our main contributions include:

\begin{itemize}
	\item We investigate energy-aware graph substitution, which allows energy saving or its tradeoff with performance to be an optimization objective. Unlike model pruning that loses accuracy for energy savings, substitution maintains accuracy critical for most if not all ML applications. 
	\item We propose a search algorithm that searches through the larger space of equivalent graphs and associated algorithm assignment for greater energy savings in a timely manner.
	\item We propose an energy model that predicts the inference time, power and energy consumption of graph, critical for online deployment and its integration to DNN frameworks. 
	\item We implement the optimization method and  evaluate it  with multiple widely used  DNNs. Our results on real systems are promising, i.e., 24\% energy savings with negligible performance impact.
\end{itemize}

\section{Related Work}
This work is closely related to research in two areas: energy-aware pruning and graph substitution based optimization.

\textbf{Energy-aware pruning. } such method reduces energy consumption through weight pruning or filter pruning. The former reduces the number of non-zero model parameters and thus reduces computation and memory footprint and required energy. The latter reduces the number of filters applied to the input data and intermediate layers of DNNs, thus reducing computation and required energy. Both are at the cost of reduced accuracy as in~\citet{yang2017designing}. 
Our method maintains accuracy and instead explores the trade-off between energy and time.
Another distinction is that our method can work on cuDNN~\cite{chetlur2014cudnn} and Nvidia GPU, which can be treated as a blackbox for model training and optimization, while existing pruning work relies on the detailed knowledge of algorithms and hardware\cite{chen2016eyeriss} and use them to guide optimization.

\textbf{Graph substitution based optimization.} Such method generates equivalent graphs and searches for the one with minimum cost.  
While existing work mainly aims to optimize time, ours investigates energy optimization and balances between energy and performance.
Some frameworks including TensorFlow, PyTorch, and TVM~\cite{chen2018tvm}  use greedy rule-based methods to transfer a graph into an optimized one. MetaFlow~\cite{jia2019opt_dnn} is most related to ours. However, our difference is multi-fold. First, MetaFlow supports time and a few other cost dimensions (e.g. FLOPS, memory usage,number of kernel) as  optimization objectives, while ours additionally  supports energy and power.
Second, theirs searches through the equivalent graphs to find a graph that minimizes the cost function, while ours finds a graph and additionally an associated algorithm assignment that minimizes the cost function. 
%


\section{Energy-aware Graph Optimization}

In this section, we first overview the optimization method and introduce a few notations to support  rigorous discussions. Then we present the details of the modeling and algorithm design that realize the optimization.

\subsection{Overview and Notations}
Similar to the existing DNN optimizer (\citet{abadi2016tensorflow}; PyTorch; \citet{chen2018tvm}; \citet{jia2019opt_dnn}), ours uses a \texttt{computation graph} to represent the computation of a DNN model.
A \texttt{computation graph} $\mathcal{G}$ consists a set of \texttt{nodes} and \texttt{edges}. Each \texttt{node} is an operator (e.g., convolution, max pooling, add) and each \texttt{edge} is a tensor. 
Sometimes we call a \texttt{computation graph} a \texttt{graph} for short.
A \texttt{graph} $\mathcal{G}$ takes one or more input tensors $\mathcal{I}$ and produces one or more output tensors $\mathcal{O}$.

Two \texttt{graphs} $\mathcal{G}$ and $\mathcal{G}'$ are considered \texttt{equivalent} if for any input tensors they produce the same output tensors.
A \texttt{graph substitution} $\mathcal{S}$ takes a computation graph $\mathcal{G}$, transforms a subgraph of $\mathcal{G}$  by using some  rules, therefor generates one (or more) new \texttt{graph} $\mathcal{G}'$. If $\mathcal{G}$ and $\mathcal{G}'$ are always \texttt{equivalent}, $\mathcal{S}$ is called an \texttt{equivalent graph substitution}.
Given a \texttt{graph} $\mathcal{G}$ and a set of \texttt{equivalent graph substitutions} $\{\mathcal{S}_i\}$, the set of all \texttt{graphs} $\{\mathcal{G}_i\}$ that can be generated by zero or more \texttt{equivalent graph substitutions} is called the \texttt{equivalent graph space} or \texttt{graph space} for short.

For a given node of a \texttt{computation graph}, there exist one or more implementations that can perform the computation of the operator\footnote{For examples, in cuDNN there are eight kernels that implement the computation of convolution at same accuracy.}. We call each implementation of the computation an \texttt{algorithm}\footnote{We will use \texttt{algorithm} when we refer to this term, we will use algorithm(with normal font) when we use the word in a normal sense.} 
of the node. 
An \texttt{algorithm assignment} $\mathcal{A}$ of a \texttt{graph} $\mathcal{G}$ maps each node of $\mathcal{G}$ to an \texttt{algorithm}. For any given \texttt{graph} $\mathcal{G}$ and  node $\mathcal{N}$, assuming we have a method\footnote{Such method could be provided by the inference engine or underlying libraries such as cuDNN.} of knowing all \texttt{algorithms} of $\mathcal{N}$,  we can know the set of possible \texttt{algorithm assignments} of $\mathcal{G}$. 
Given two \texttt{algorithm assignments} $\mathcal{A}_1$ and $\mathcal{A}_2$ of $\mathcal{G}$, \texttt{distance}$(\mathcal{A}_1,\mathcal{A}_2)$ is defined as the number of nodes  being mapped to different \texttt{algorithms}.

A \texttt{cost function} takes a \texttt{graph} $\mathcal{G}$ and associated \texttt{algorithm assignment} $\mathcal{A}$ as input, and outputs a quantitative value as  \texttt{cost}. Here \texttt{cost} is typically specified by users and examples include inference time, inference energy and inference power.


Essentially, the core of our energy-aware DNN graph optimization is a search algorithm.
Given a \texttt{cost function} $Cost()$ defined by user, the algorithm  takes a \texttt{graph} $\mathcal{G}$ and a set $\{S_i\}$ of \texttt{equivalent graph substitutions} as input, returns a equivalent \texttt{graph} $\mathcal{G}_{opt}$ and an associated \texttt{algorithm assignment} $\mathcal{A}_{opt}$, so that  $Cost(\mathcal{G}_{opt},\mathcal{A}_{opt})$ is the minimum.
With the obtained $(\mathcal{G}_{opt},\mathcal{A}_{opt})$, we can run the \texttt{graph} $\mathcal{G}_{opt}$ on existing inference engines using the \texttt{algorithm assignment} $\mathcal{A}_{opt}$.

  
 







\subsection{Cost Model}

Our optimizer allows users to specify a \texttt{cost function} that can use a single metric or combine multiple metrics as optimization objectives.
Although the \texttt{cost function} supports many  cost metrics (e.g. FLOPS, memory usage), in this paper we focus on energy, inference time, and derived metric such as average power.
Given a \texttt{graph} $\mathcal{G}$ and \texttt{algorithm assignment} $\mathcal{A}$,  energy, inference time and power of $(\mathcal{G},\mathcal{A})$ are denoted  as $Energ_{(\mathcal{G},\mathcal{A})}$, $Time_{(\mathcal{G},\mathcal{A})}$ and $Power_{(\mathcal{G},\mathcal{A})}$.

Depending on preference, users can define a \texttt{cost function} on their own, or use one of the following, where $0\le w \le 1$ specifies the weight on energy:
$$Cost(G,A)=w*Energ_{(G,A)}+(1-w)*Time_{(G,A)} $$ which is a linear function of energy and time,
$$Cost(G,A)=Energ_{(G,A)}^w*Time_{(G,A)}^{(1-w)}$$ which is a product function of energy and time, and 
$$Cost(G,A)=Power_{(G,A)}$$ which is the energy-to-time ratio.

Given any $\mathcal{G}$ and $\mathcal{A}$, the most straightforward way to evaluate $Energ_{(\mathcal{G},\mathcal{A})}$,$Time_{(\mathcal{G},\mathcal{A})}$ and $Power_{(\mathcal{G},\mathcal{A})}$ is to measure them directly.
But this method is infeasible because the number of $(\mathcal{G},\mathcal{A})$ in the search space is too large.
To alleviate this challenge, we build \texttt{cost models} to predict the costs without actually running the graphs and measuring them. 
An observation is, although the search space is large, the number of nodes with different parameters is much smaller.
For each node $\mathcal{N}$ of a \texttt{graph} we measure its power consumption and inference time under each available \texttt{algorithm}, and use the model below to calculate the energy, inference time and power of ($\mathcal{G}$,$\mathcal{A}$):
$$ Energ_{(\mathcal{G,A})} = \sum\limits_{n \in \mathcal{G}.nodes} {Energ_{(n,\mathcal{A}(n))}} $$
$$ Tim{e_{(\mathcal{G,A})}} = \sum\limits_{n \in \mathcal{G}.nodes} {Tim{e_{(n,\mathcal{A}(n))}}} $$
\[Powe{r_{(\mathcal{G,A})}} = \frac{{Energ_{(\mathcal{G,A})}}}{{Tim{e_{(\mathcal{G,A})}}}}\]
The energy consumption of a \texttt{graph} with an \texttt{algorithm assignment} is the sum of energy over all the  nodes under the corresponding \texttt{algorithms}, and so is the inference time.
The power consumption is energy divided by time.
With these models,  nodes (even for different graphs) with the same parameters only need to be measured once. The measured values are stored in a database and persisted onto disk for future lookup.
Our cost model was inspired by \citet{jia2018exploring}'s cost model of inference time. We extend it to support power and energy, and add \texttt{algorithm assignment} as a new input dimension.

\subsection{The Search Algorithm}

As stated, the core of our energy-aware optimizer is a search algorithm that identifies a equivalent \texttt{graph} $\mathcal{G}_{opt}$ and an associated \texttt{algorithm assignment} $\mathcal{A}_{opt}$ that minimizes the cost.  This search algorithm has an outer level and  an inner level. 
 
The outer level searches for an optimized equivalent \texttt{graph} $\mathcal{G}'$ of $\mathcal{G}$ in the \texttt{graph space} as shown in algorithm \ref{outsearch}.
Similar to the relaxed search algorithm proposed by \citet{jia2019opt_dnn}, this algorithm uses a parameter $\alpha$ to tradeoff between the search time
and the best-discovered solution. The parameter has the same meaning as the one in \citet{jia2019opt_dnn}'s paper, where with $\alpha = 1$ the algorithm becomes a simple greedy algorithm, and as $\alpha$ increases, the search algorithm explores a larger part of the search space.

The inner level searches for an optimized \texttt{algorithm assignment} $\mathcal{A}$ for a given graph $\mathcal{G}$ as shown in algorithm \ref{insearch}.
For a given \texttt{graph} $\mathcal{G}$, it first picks up an arbitrary \texttt{algorithm assignment} $\mathcal{A}$ as the start point. Then it searches for a different \texttt{algorithm assignment} with a better cost in the neighborhood of $\mathcal{A}$ with \texttt{distance} $d$. If found, it updates the $\mathcal{A}$ with the found \texttt{algorithm assignment}, then repeats the search process. The algorithm terminates when no better \texttt{algorithm assignment} can be found any more. Here the parameter $d$ can be set by users. If $d=1$, the inner search is a simple greedy algorithm. If $d=2$, the inner search is still greedy but allows one step of downgrade for the search goal. When $d$ is larger than the number of node of $\mathcal{G}$, the search becomes exhaustive. 

Essentially our inner algorithm is also a relaxed greedy search but  in a different perspective than \citet{jia2019opt_dnn}'s method.
Although the algorithm in~\citet{jia2019opt_dnn} performs well in searching equivalent graphs, it is problematic for  searching  algorithm assignment for some \texttt{cost functions} such as energy. With their algorithm, even a relaxed value $\alpha=1.01$ can lead to a significant increase in searching time.  
 

It is easy to verify for any \texttt{cost function} that is a linear combination of inference time and energy, the inner search with $d=1$ is sufficient to find an optimal \texttt{algorithm assignment}. Because with the \texttt{cost function} as a linear combination of time and energy, the \texttt{cost function} is also a linear combination of time and energy of all nodes. 

\begin{algorithm}[h]
	\caption{Outer Search Algorithm}
	\label{outsearch}
	\textbf{Premise:} A cost model $Cost()$ and a parameter $\alpha$.
	
	\textbf{Input:} An initial computation graph
	$\mathcal{G}_0$, a set of equivalent graph substitutions $\{ S_1,...,S_m \}$. 
	
	\textbf{Output:} An optimized computation graph with an associated algorithm assignment: $(\mathcal{G}_{opt}$, $\mathcal{A}_{opt})$.
	
	\begin{algorithmic}[1]
		\STATE $\mathcal{A}_0 = innerSearch(\mathcal{G}_0) $;
		\STATE $Q$ = $\left \{ (\mathcal{G}_0,\mathcal{A}_0) \right \}$;
		\STATE $(\mathcal{G}_{opt},\mathcal{A}_{opt})=(\mathcal{G}_0,\mathcal{A}_0)$;
		\WHILE{ $Q$ != $\left \{ \right\}$} 
		\STATE $(\mathcal{G},\mathcal{A}) = Q.dequeue()$ 
		\FOR{$\mathcal{G}' \in S_i(\mathcal{G}), i\in 1..m $}
		\STATE $\mathcal{A}'= innerSearch(\mathcal{G}')$;
		\IF{ $Cost(\mathcal{G}',\mathcal{A}') < Cost(\mathcal{G}_{opt},\mathcal{A}_{opt}) $}
		\STATE ($\mathcal{G}_{opt},\mathcal{A}_{opt}) = (\mathcal{G}', \mathcal{A}')$
		\ENDIF
		\IF{ $Cost(\mathcal{G}',\mathcal{A}') < \alpha* Cost(\mathcal{G}_{opt},\mathcal{A}_{opt}) $}
		\STATE $Q.enqueue(\mathcal{G}',\mathcal{A}')$
		\ENDIF
		\ENDFOR
		\ENDWHILE
		\STATE \textbf{return} $(\mathcal{G}_{opt},\mathcal{A}_{opt})$
	\end{algorithmic}
\end{algorithm}

\begin{algorithm}
	\caption{Inner Search Algorithm}
	\label{insearch}
	
	\textbf{Premise:} A cost model $Cost()$ and a parameter $d$.
	
	\textbf{Input:} A computation graph $\mathcal{G}$.
	
	\textbf{Output:} An optimized \texttt{algorithm assignment} $\mathcal{A}$ of $\mathcal{G}$.
	\begin{algorithmic}[1]
		\STATE Let $\mathcal{S}$ be the set of all \texttt{algorithm assignments} of $\mathcal{G}$
		\STATE Pick $\mathcal{A} \in \mathcal{S}$ arbitrarily.
		\REPEAT
		\STATE Initialize $noChange=true$
		\FOR {$ \mathcal{A}' \in \{\mathcal{A}'|\mathcal{A}'\in \mathcal{S} \wedge distance(\mathcal{A}',\mathcal{A})<=d\}$}
		\IF {$Cost(\mathcal{G},\mathcal{A}') < Cost(\mathcal{G},\mathcal{A})$}
		\STATE $\mathcal{A}= \mathcal{A}'$
		\STATE $noChange = false$
		\ENDIF
		\ENDFOR
		\UNTIL $noChange$ is $true$
		\STATE \textbf{return} $\mathcal{A}$
	\end{algorithmic}
\end{algorithm}



\subsection{Implementation}

Since we use the backtracking search proposed by \citet{jia2019opt_dnn} as our outer search, we build our implementation based on MetaFlow\cite{jia2019opt_dnn}. Specifically, we modify the code of the original optimizer of MetaFlow to our energy-aware optimizer. We also make some changes so that our optimized \texttt{graph} $\mathcal{G}$ and associated \texttt{algorithm assignment} $\mathcal{A}$ can be run on the MetaFlow's built-in inference engine.

\section{Evaluation}

In this section,  we first introduce the experiment setup, then present the evaluation results in multiple aspects. The evaluations focus on CNN graphs in this paper.



\subsection{Experimental Setup}

The DNN models we use for evaluation include Inception-v3~\cite{szegedy2016rethinking}, SqueezeNet~\cite{iandola2016squeezenet} and ResNet-50~\cite{he2016deep}. They are the state-of-the-art convolution DNN models. 

In our experiments we use the build-in inference engine of MetaFlow~\cite{jia2019opt_dnn}. The reason we use this engine is two-fold.  First, according to their paper, it outperforms TensorFlow, TensorRT~\cite{tensorrt2019programmable} and TensorFlowXLA~\cite{leary2017xla}, and can be considered the state-of-the-art research. Second, our implementation is built on top of theirs, it is easy to run the optimized $(\mathcal{G},\mathcal{A})$ on their build-in engine without the need of exporting and importing graphs.

Our computer system has  two 20-cores Intel Xeon Gold 6148 CPUs and two Nvidia Tesla V100 GPUs. In all experiments we only use one CPU core and one V100 GPU.

We set $\alpha=1.05$ for the outer search,  $d=1$ for the inner search for linear combination of time and energy, $d=2$ otherwise for other optimization objectives.

For each DNN model, the first run of our optimizer takes longer, since we need to profile the inference time, energy, and power for \texttt{cost model}. After the first run, each later run finishes in a few minutes since most profile results needed by our \texttt{cost model} have already been cached into database.

To measure energy consumption of a \texttt{graph}, we use the tool \texttt{nvidia-smi} to monitor the real-time power of a running \texttt{graph}. We sample the power consumption periodically, then we multiply the average power consumption with the inference time to get energy consumption of a \texttt{graph}. To get a reliable result,we run a \texttt{graph} for 4 seconds before sampling with \texttt{nvidia-smi}, and measure for at least another 4 seconds. Inference time is measured at same time as power, the value is calculated from how many inference is done in the measure period.

\begin{table*}[t]
	\caption{ Accuracy of Cost Model (SqueezeNet)}
	\label{table2}
	\begin{center}
		\begin{small}
			\begin{sc}
				\begin{tabular}{c|c|cccccccc}
					\hline
					& & Graph1 & Graph2 & Graph3 & Graph4 & Graph5 & Graph6 & Graph7 & Graph8 \\
					\hline
					time & estimated & 0.862 & 0.801 & 0.778 & 0.758 & 0.740 & 0.721 & 0.696 & 0.686 \\
					& actual& 0.895 & 0.833 & 0.802 & 0.805 & 0.795 & 0.796 & 0.758 & 0.732 \\
					\hline
					power & estimated & 80.78 &83.28 & 84.10 &84.77 & 85.49 & 86.53 & 89.20 & 90.45 \\
					& actual &79.00 &81.50 & 81.83 & 81.83 &82.00 & 82.33 & 85.00 & 87.00 \\ 
					\hline
					energy & estimated & 69.66 & 66.73 & 65.42 & 64.33 & 63.32 & 62.43 &62.13 & 62.12 \\
					& actual & 70.70 & 67.88 & 65.62 & 65.87 & 65.21 & 65.54 & 64.49 &63.75 \\
					\hline
				\end{tabular}
			\end{sc}
		\end{small}
	\end{center}
	\vskip -0.1in
\end{table*}

\begin{table*}[t]
	\caption{Various Goals on 3 CNN Graphs}
	\label{table1_5}
	\begin{center}
		\begin{small}
			\begin{sc}
				\setlength{\tabcolsep}{1.5mm}{}
				\begin{tabular}{l|c|ccc|ccc|ccc}
					\hline
					&&\multicolumn{3}{|c|}{Squeezenet} &\multicolumn{3}{|c|}{InceptionV3}&\multicolumn{3}{c}{ResNet} \\
					&graph&time&power&energy&time&power&energy&time&power&energy \\
					\hline
					
					&origin &0.916&101.2&92.72&6.796&71.16&483.6&2.079&79.55&165.4 \\
					&MetaFLow best time &\textbf{0.749}&112.1&84.04&\textbf{5.493}&83.00&455.9&\textbf{2.033}&80.00&162.6 \\
					\hline
					Our Work&best time &\textbf{\color{red}0.683}&107.2&73.26&\textbf{\color{red}5.469}&82.00&448.4&\textbf{\color{red}1.926}&78.00&150.2 \\
					&best energy &0.736&86.50&\textbf{\color{red}63.67}&5.788&72.13&\textbf{\color{red}417.5}&1.982&75.00&\textbf{\color{red}148.6}\\
					&best power &2.183&\textbf{\color{red}59.20}&129.2&12.59&\textbf{\color{red}53.25}&670.6&6.540&\textbf{\color{red}52.27}&341.9 \\
					&0.5power+0.5energy &0.935&\textbf{75.40}&\textbf{70.52}&7.742&\textbf{59.33}&\textbf{459.3}&2.233&\textbf{67.81}&\textbf{151.4} \\
					\bottomrule
				\end{tabular}
			\end{sc}
		\end{small}
	\end{center}
	\vskip -0.1in
\end{table*}

\begin{table}[t]
	\caption{Balance between Time and Energy (SqueezeNet)}
	\label{table1_7}
	\begin{center}
		\begin{small}
			\begin{sc}
				\begin{tabular}{l|ccc}
					\hline
					graph&time&power&energy\\
					\hline
					best time &0.683&107.2&73.26 \\
					0.8time+0.2energy &0.693&101.2&70.16\\
					0.6time+0.4energy &0.707&93.50&66.13\\
					0.4time+0.6energy &0.719&89.00&64.06\\
					0.2time+0.8energy &0.715&89.33&63.88\\
					best energy &0.736&86.50&63.67\\
					\hline
				\end{tabular}
			\end{sc}
		\end{small}
	\end{center}
	\vskip -0.1in
\end{table}
\begin{table}[t]
	\caption{Contribution of Inner Search  (SqueezeNet)}
	\label{table1_8}
	\begin{center}
		\begin{small}
			\begin{sc}
				\begin{tabular}{l|ccc}
					\hline
					graph&time&power&energy\\
					\hline
					origin &0.916&101&92.72 \\
					outer search only&0.776&108&83.87\\
					inner search only&0.843&91.8&77.40\\
					both inner and outer&0.736&86.5&63.67\\
					\hline
				\end{tabular}
			\end{sc}
		\end{small}
	\end{center}
	\vskip -0.1in
\end{table}

\subsection{Accuracy of Cost Model}
In this section, we use several \texttt{graphs} (with associated \texttt{algorithm assignments}) from the search process of SqueezeNet as examples to evaluate the model accuracy. We  measure the actual time, energy and power consumption and compare them with the model estimations.
Table \ref{table2} shows although the estimated values and the actual ones can be different by up to 10\%, it correctly projects the orders of the assignments, and identifies the one with the minimum cost, which is the ultimate goal. 

\subsection{Evaluation on Various Objectives}
In this section, we evaluate our optimized results of three state-of-the-art DNNs with different optimization objectives, and compare them with a few counterparts.
The results are in Table \ref{table1_5}.
``Origin'' stands for the original un-optimized \texttt{graph}, ``MetaFlow Best Time'' stands for the \texttt{graph} optimized by MetaFlow's optimizer toward best inference time. The rest are the \texttt{graphs} optimized by our optimizer. ``Best Time'' stands for the \texttt{cost function} is $Cost(G,A)=Time_{(G,A)} $, ``0.5Power+0.5Energy'' stands for the \texttt{cost function} is $Cost(G,A)=0.5{\times}Power{(G,A)}+0.5{\times}Energ_{(G,A)}$ etc.

\textbf{Energy:} When  energy is the optimization objective, on SqueezeNet,  our optimized \texttt{graph} consumes  24\% less energy than MetaFlow optimized. In addition, our inference time is slightly shorter. On Inception-v3, ours consumes 8\% less energy at the cost of 5\% more inference time. On ResNet, ours consumes 9\% less energy \textit{and} is faster than MetaFlow optimized by 2\%.

\textbf{Power:} When minimum power is the objective, our optimized \texttt{graphs} use 47\%, 36\% and 35\% less power than MetaFlow optimized and 40\%, 25\% and 35\% less than origin, although at the cost of more inference time and energy. This kind of optimization might be helpful for battery-powered devices with desired functioning duration. When we balance between power and energy, our optimized graphs use 33\%, 29\% and 16\% less power than MetaFlow optimized \texttt{graphs}, and 24\%, 17\% and 16\% less than origin, at a much more acceptable cost of inference time and energy.

\textbf{Time:} When minimum inference time is the objective, ours outperforms MetaFlow by 9\% on SqueezeNet, 5\% on ResNet, and performs similarly  on Inception-v3. The significant improvement of inference time is a pleasant surprise to us as we only set out to support energy awareness. We attribute this improvement to better algorithm assignment, which is enabled by our accurate profiling of \texttt{graph} nodes and cost estimations. 


\subsection{Tradeoff between Time and Energy}
In this section, we illustrate our optimizer's ability of balancing between multiple metrics, especially between time and energy.
We use  \texttt{cost function} $Cost(G,A)=w{\times}Time_{(G,A)}+(1-w){\times}Energ_{(G,A)}$, and change $w$ from 1 to 0. When $w$=1, the optimization objective is the best inference time. We use normalized values of time and energy
in the \texttt{cost function}, so that the weight $w$ makes better sense.
The results in Table \ref{table1_7} show that our algorithm is able to get a smooth balance between inference time and energy. This means users are able to balance inference time and energy at their preference.

This kind of balance allows many possibilities. For example,  from Table \ref{table1_7} we know the lower bound of inference time of Squeezenet is 0.683ms, and energy per 1000 inference is 63.67J. Optimization objectives with hard restrictions such as ``less energy as possible, while inference time is faster than 0.7ms'' can be easily supported by binary searching on weight $w$. By only requiring pair-wise accuracy, such binary search is more accurate than the one suggested by \citet{jia2019opt_dnn}, which relies heavily on the value accuracy of cost models.

\subsection{Importance of Inner Search}
In this section we show the importance of introducing the inner search.
We use SqueezeNet as the study case, and set energy as the optimization objective. We compare the performance under several algorithm configurations: turn off the inner and outer searches, turn on the outer search only,  turn on the inner search only, and turn on both. 


Table \ref{table1_8} shows, with both searches enabled, energy can be reduced by 31\% compared to the configuration where both are disabled (denoted as origin).  With the outer search only, energy is only reduced by 10\%. With the inner search only, energy is reduced by 16\%.  This shows the inner search plays a significant role in energy-aware graph optimization.

\section{Conclusion and Future Work}
We investigate energy-aware graph substitution and propose a cost model and search algorithm to support it. We implement an energy-aware graph optimizer that allows optimization for time, power, energy and the tradeoffs between them. We evaluate our work and show promising results on multiple commonly used DNNs.

In the future we plan to evaluate our methods with  more types of DNNs and inference engines. We will also investigate to introduce accuracy into our cost model and search algorithm, and support the tradeoffs between accuracy and other metrics. 





\bibliography{example_paper}
\bibliographystyle{sysml2019}

\appendix
\section{Source Code}
\url{https://github.com/wangyu-/mlsys20_workshop}
%


\end{document}